\documentclass[twocolumn]{autart}    

\usepackage{graphicx}          
\usepackage[dvips]{epsfig}    

\usepackage{cite}

\usepackage{textcomp}
\usepackage{amsmath,amsfonts,amssymb,color}

\def\endpf{\hspace*{\fill}~$\square$\par\endtrivlist\unskip}

\usepackage{epsfig}
\usepackage{psfrag}
\usepackage{algorithm}
\usepackage{algpseudocode}
\usepackage{array}
\usepackage{epstopdf}
\usepackage{mathtools}
\usepackage{enumerate}
\usepackage{soul,color}
\usepackage{wrapfig}
\usepackage{nccmath}

\newtheorem{lemma}{Lemma}
\newtheorem{remark}{Remark}
\newtheorem{definition}{Definition}

\newtheorem{assumption}{Assumption}

\DeclareMathOperator{\tr}{tr}

\begin{document}

\begin{frontmatter}

\title{Online Change Points Detection for Linear Dynamical Systems with Finite Sample Guarantees\thanksref{footnoteinfo}} 

\thanks[footnoteinfo]{This work was supported by the National Science Foundation CAREER award 1653648. This work represents the opinions of the authors and not the USDA or NIFA. }

\author[EE]{Lei Xin}\ead{lxin@purdue.edu},    
\author[ME]{George Chiu}\ead{jgchiu@purdue.edu},               
\author[EE]{Shreyas Sundaram}\ead{sundara2\@purdue.edu}  

\address[EE]{Elmore Family School of Electrical and Computer Engineering, Purdue University, West Lafayette, IN 47907, USA}  
\address[ME]{School of Mechanical Engineering, Purdue University, West Lafayette, IN 47907, USA}             


\begin{abstract}                          
The problem of online change point detection is to detect  abrupt changes in properties of time series, ideally as soon as possible after those changes occur. Existing work on online change point detection either assumes i.i.d data, focuses on asymptotic analysis, does not present theoretical guarantees on the trade-off between detection accuracy and detection delay, or is only suitable for detecting single change points. In this work, we study the online change point detection problem for linear dynamical systems with unknown dynamics, where the data exhibits temporal correlations and the system could have multiple change points. We develop a data-dependent threshold that can be used in our test that allows one to achieve a pre-specified upper bound on the probability of making a false alarm. We further provide a finite-sample-based bound for the probability of detecting a change point.  Our bound demonstrates how parameters used in our algorithm affect the detection probability and delay, and provides guidance on the minimum required time between changes to guarantee detection. 
\end{abstract}

\end{frontmatter}

\section{Introduction}

The problem of change point detection (CPD) is to detect when abrupt changes in properties of time series occur. This problem has wide application in various fields, including monitoring of medical conditions, speech recognition, environmental surveillance, and image analysis \cite{aminikhanghahi2017survey}. In an offline setting, the goal is to determine the change points by looking at the entire dataset all at once \cite{truong2020selective}. In contrast, in an online (or sequential) setting, the goal is to detect when changes occur as soon as possible based on new data points arriving in a streaming manner  \cite{polunchenko2012state}.  

Online CPD under the assumption of independence of samples over time has been studied extensively over the past few decades \cite{tartakovsky2012efficient, flynn2019change, chang2019kernel, li2015m}, and many approaches have been proposed; for example,  \cite{shiryaev1963optimum,adams2007bayesian} leveraged Bayesian methods for online CPD, while \cite{titsias2022sequential,lee2022training} explored the use of neural networks.  The CPD problem is more challenging when the data exhibits correlations over time, and coming up with theoretical guarantees for such settings is an active area of research \cite{tartakovsky2017asymptotic}. Indeed, time-correlations between data are commonly observed in practice.  For example, data generated by  dynamical systems is inherently correlated over time; such systems commonly occur in control theory, machine learning and economics \cite{zadeh2008linear}. As a practical example, when learning system dynamics from observed data, one may want to know if the system changes dynamics at some point of time to avoid using biased data \cite{xin2022identifying}. There are several papers that focus on CPD for dynamical systems \cite{zoeter2005change, kawahara2007change, fox2008nonparametric}. However, the existing works typically do not provide finite sample guarantees for the probabilities of making false and true alarms, i.e., it is unclear how well these algorithms perform given a finite number of data points. 

A common approach used for online CPD is to compare a data-based statistic against a given threshold, with an alarm being raised if the statistic is larger than the threshold. There are some papers that provide threshold values that can achieve desired probabilities of false and true alarms, based on Bayesian approaches \cite{lai1998information, han2019confirmatory}.  However, these works typically require a known probability distribution of the data prior to the change or do not have a theoretical analysis that demonstrates the relationship between detection accuracy and detection delay. Indeed, a rigorous characterization of the relationship between detection accuracy and detection delay is rarely provided in the literature. The paper \cite{alanqary2021change} studies change point detection leveraging multivariate singular spectrum analysis, where the authors model the time series data as being generated by a spatio-temporal model and theoretically demonstrate the trade-off between detection delay and false alarm rate. However, the main focus of their analysis is on how a user-specified threshold value used in their test affects the performance of the algorithm in expectation. In contrast, our focus in this paper is on finite sample guarantees for the probabilities of making false and true alarms, and how the number of samples used in our algorithm at each time step affects the detection accuracy and detection delay.

Lastly, we note that all of the above mentioned works do not theoretically demonstrate how the presence of multiple change points affects the performance of the CPD algorithms. This aspect is captured in our algorithm and its associated analysis. 

In summary, our contributions are as follows. 
\begin{itemize}
  \item We provide an online change point detection algorithm for linear dynamical systems that is suitable for multiple change points. The algorithm is based on a least squares approach and can be easily implemented. We develop a data-dependent threshold that can be used in our test, which enables the user to achieve a pre-specified false alarm probability (assuming certain prior knowledge about the system is available). We note that the threshold does not require perfect knowledge of the system parameters at any given time. This is different from many existing works on CPD, e.g., \cite{lai1998information,geng2019online, banerjee2015data}, which usually require known distribution prior to the change (when there is only a single change point).  
 
  \item We provide a finite-sample-based lower bound for the probability of making a true alarm after changes occur with a certain delay. Our result demonstrates the trade-off between accurate detection and detection delay, as well as the ability to detect fast changes. 
  \end{itemize}

Our theoretical analysis and guarantees are demonstrated and validated via numerical examples at the end of the paper.

\section{Mathematical Notation and Terminology}
Let $\mathbb{R}$ and $\mathbb{N}$ denote the set of real numbers and natural numbers, respectively. The symbols $\cup$ and $\cap$ are used to denote the union and intersection of sets, respectively. Vectors are treated as column vectors, and the symbol $'$ is used to denote the transpose operator.  Let $\lambda_{min}(\cdot)$ be the smallest eigenvalue of a symmetric matrix. The spectral radius of a given matrix is denoted as $\rho(\cdot)$. We use $\|\cdot\|$ and $\|\cdot\|_{F}$ to denote the spectral norm and Frobenius norm of a given matrix, respectively. We use $\tr(\cdot)$ to denote the trace of a given matrix. A Gaussian distributed random vector is denoted as $u\sim \mathcal{N}(\mu,\Sigma)$, where $\mu$ is the mean and $\Sigma$ is the covariance matrix. We use $I_{n}$ to denote the identity matrix with dimension $n \times n$. The symbol $\sigma(\cdot)$ is used to denote the sigma field generated by the corresponding random vectors. We use $\mathcal{S}^{n-1}$ to denote the unit sphere in $n$-dimensional space.
\section{Problem formulation and algorithm}
Consider the  discrete time linear dynamical system
\begin{equation} \label{system}
\begin{aligned} 
   x_{k+1}&=A_{k}x_{k}+B_{k}u_{k}+w_{k},
\end{aligned}
\end{equation}
where $k\in \mathbb{N}$ is the time index, $x_{k}\in \mathbb{R}^{n}$ is the state, $u_{k}\in \mathbb{R}^{p}$ is the input, $w_{k}\in \mathbb{R}^{n}$ is the process noise, $A_{k}\in \mathbb{R}^{n \times n}$ and $B_{k}\in \mathbb{R}^{n \times p}$ are system matrices. The input and process noise are assumed to be i.i.d Gaussian, with $u_{k} \sim \mathcal{N}(0,\sigma_{u}^{2}I_{p})$ and $w_{k} \sim \mathcal{N}(0,\sigma_{w}^{2}I_{n})$. We define $\sigma_{min}\triangleq\min(\sigma_{w},\sigma_{u})$. 
The initial state $x_{0}$ is assumed to be independent of $u_{k}$ and $w_{k}$. The system matrices $A_{k},B_{k}$ are deterministic but unknown. Let $\Theta_{k}=\begin{bmatrix} A_{k} &B_{k}\end{bmatrix}$. We further assume that there are known parameters $b_{\sigma_{w}}$ and $b_{\Theta}$ that satisfy $\sigma_{w}\leq b_{\sigma_{w}}$ and $\|\Theta_{k}\|\leq b_{\Theta}$ for all $k\geq 0$. We call a time index $\hat{k}\in \mathbb{N}_{\ge 1}$ a {\it change point} if  $\Delta_{\hat{k}}\triangleq\|\Theta_{\hat{k}-1}-\Theta_{\hat{k}}\|>0$. Our goal is to determine the change points using observed data from system \eqref{system} in an online fashion. From the perspective of control theory, one can treat the above model as a switched system with unknown dynamics \cite{sun2005analysis}, where the goal is to detect when the system switches dynamics and learn a model for each mode of the system, so that one can design a better controller; one can also treat $A_{k}$ in system \eqref{system} as a closed loop system under control, and the input $u_{k}$ as a small exploratory input (which is a common strategy used in the adaptive control literature \cite{dean2018regret}).


Let $N\geq 2$ be a design parameter that corresponds to the length of the interval of previously seen data we would like to use at each step to detect change points (our analysis later will provide guidance on how to select $N$). Let $\{(x_{i},u_{j}):0 \leq i \leq 2N-1,0 \leq j \leq 2N-2\}$ denote the initial dataset. At each time step $k\geq 2N-1$, a sample $(x_{k+1},u_{k})$ generated by system \eqref{system} is observed. Let the label matrices for the {\it reference window} and the {\it test window} at time step $k$ be 
\begin{equation} 
\begin{aligned}
X^{ref}_{k}=\begin{bmatrix}
x_{k-2N+3}&\cdots&x_{k-N+1}
\end{bmatrix}  \in \mathbb{R}^{n\times (N-1)},\\
\end{aligned}
\end{equation}
and 
\begin{equation} 
\begin{aligned}
X^{test}_{k}=\begin{bmatrix}
x_{k-N+3}&\cdots&x_{k+1}
\end{bmatrix}  \in \mathbb{R}^{n\times (N-1)},
\end{aligned}
\end{equation}
respectively. Let $z_{k}=\begin{bmatrix} x_{k}'& u_{k}'\end{bmatrix}^{'}\in \mathbb{R}^{n+p}$ for $k\geq 0$, and define the regressor matrices for the reference window and the test window at time step $k$ as 
\begin{equation} 
\begin{aligned}
&Z^{ref}_{k}=\begin{bmatrix}
z_{k-2N+2}&\cdots&z_{k-N}
\end{bmatrix} \in \mathbb{R}^{(n+p)\times (N-1)},\\
&Z^{test}_{k}=\begin{bmatrix}
z_{k-N+2}&\cdots&z_{k}
\end{bmatrix} \in \mathbb{R}^{(n+p)\times (N-1)}.
\end{aligned}
\end{equation}
Furthermore, we denote the noise matrices for the reference window and the test window at time step $k$
as
\begin{equation} 
\begin{aligned}
&W^{ref}_{k}=\begin{bmatrix}
w_{k-2N+2}&\cdots&w_{k-N}
\end{bmatrix} \in \mathbb{R}^{n\times (N-1)},\\
&W^{test}_{k}=\begin{bmatrix}
w_{k-N+2}&\cdots&w_{k}
\end{bmatrix} \in \mathbb{R}^{n\times (N-1)}.
\end{aligned}
\end{equation}

Based on these data windows, we consider the following intuitive approach. Let $\hat{\Theta}_{k}^{ref}$ and $\hat{\Theta}_{k}^{test}$ be the estimated system models using the data from the reference window and the test window, respectively. If the data from the reference window and the test window are generated by the same dynamics, then their estimated models should be similar; if the data from these windows are generated by different dynamics, i.e., if there is a change point, their estimated models should be quite different, i.e., one may flag a change point if the metric $\|\hat{\Theta}_{k}^{ref}-\hat{\Theta}_{k}^{test}\|$ is larger than some user-specified threshold value. To find $\hat{\Theta}_{k}^{ref}$ and $\hat{\Theta}_{k}^{test}$, we will solve the following regularized least squares problems at each time step $k$
\begin{equation*}
\begin{aligned}
  \mathop{\min}_{\tilde{\Theta}^{ref}_{k}\in \mathbb{R}^{n\times (n+p)}} \{\|X^{ref}_{k}-\tilde{\Theta}^{ref}_{k}Z^{ref}_{k}\|^{2}_{F}+\lambda \|\tilde{\Theta}^{ref}_{k}\|^2_{F}\},
\end{aligned}
\end{equation*}
and
\begin{equation*}
\begin{aligned}
\mathop{\min}_{\tilde{\Theta}^{test}_{k}\in \mathbb{R}^{n\times (n+p)}} \{\|X^{test}_{k}-\tilde{\Theta}^{test}_{k}Z^{test}_{k}\|^{2}_{F}+\lambda \|\tilde{\Theta}^{test}_{k}\|^2_{F}\},
\end{aligned}
\end{equation*}
where $\lambda>0$ is a regularization parameter.
The closed-form solutions of the above problems are given by
\begin{equation} \label{ref}
\begin{aligned}
\hat{\Theta}_{k}^{ref}=X_{k}^{ref}(Z_{k}^{ref})^{'}(Z_{k}^{ref}(Z_{k}^{ref})^{'}+\lambda I_{n+p})^{-1},
\end{aligned}
\end{equation}
and 
\begin{equation}  \label{test}
\begin{aligned}
\hat{\Theta}_{k}^{test}=X_{k}^{test}(Z_{k}^{test})^{'}(Z_{k}^{test}(Z_{k}^{test})^{'}+\lambda I_{n+p})^{-1},
\end{aligned}
\end{equation}
respectively. In this paper, we will develop a data-dependent threshold $\gamma_{k}=\gamma_{k}(\delta)$ that has provable finite-sample guarantees, where $\delta\in (0,1)$ is a user-specified upper bound of the false alarm probability. The specific expression of $\gamma_{k}$ will be given later when we present our results (Lemma \ref{sysid}). 

Our guarantees will apply to change points that are sufficiently separated (in time) from other change points. We make the following definition to make this formal.

\begin{definition} [Sufficiently Separated Change Point]\label{ss}
Suppose the system has $q\in \mathbb{N}_{\geq 1}$ change points. Let the sequence of change points be $0<k_{1}<\cdots<k_{q}$. We call a change point $k_{i}$, $1 \leq i\leq q-1$,  {\it sufficiently separated} if $k_{i}-k_{i-1}\geq 4N-1$ and $k_{i+1}-k_{i}\geq 4N-1$, where $k_{0}=0$.  The change point $k_{q}$ is sufficiently separated if $k_{q}-k_{q-1}\geq 4N-1$.

If the system has infinitely many change points $0<k_{1}<k_{2}<\cdots$, we call a change point $k_{i}$, $i\geq 1$, sufficiently separated if $k_{i}-k_{i-1}\geq 4N-1$ and $k_{i+1}-k_{i}\geq 4N-1$, where $k_{0}=0$.
\end{definition}


We let the value $S_{k}$ represent the most recent change point predicted by the algorithm at time step $k\geq 2N-1$, where the initialization is given by $S_{2N-2}=0$.  If the current metric  $\|\hat{\Theta}_{k}^{ref}-\hat{\Theta}_{k}^{test}\|$ is greater than the threshold $\gamma_{k}$, and $k-S_{k-1}>2N-2$, then the algorithm will predict a change point and set $S_{k}=k$; otherwise, no change point will be predicted and the algorithm will set $S_{k}=S_{k-1}$. The requirement of $k-S_{k-1}>2N-2$ is needed in addition to $\|\hat{\Theta}_{k}^{ref}-\hat{\Theta}_{k}^{test}\|\geq \gamma_{k}$ to deal with potentially multiple change points. Intuitively, if the metric is larger than the threshold for some consecutive time steps, we may not want to flag all of them as change points if all of the change points are sufficiently separated, so we  wait for a period of time to make sure that the current dynamics have settled before making the next prediction. In general, if there is no change point at time step $k$, i.e., $\|\Delta_{k}\|=0$, we want the algorithm to output $S_{k}\neq k$; if there is a change point at time step $k$, we would like the algorithm to detect that as soon as possible, i.e., if $\|\Delta_{k}\|>0$, we want the algorithm to output $S_{t}\geq k$ for some $t\geq k$, and the value $\min_{S_{t}\geq k}\{S_{t}\}$ should be small.

The above steps are encapsulated in Algorithm \ref{alg}.
\begin{algorithm}[H] 
\caption{Online Change Point Detection} \label{alg}
\textbf{Input} False alarm probability $\delta\in (0,1)$, window size $N\geq 2$, parameters $\lambda>0, b_{\sigma_{w}}, b_{\Theta}$
\begin{algorithmic}[1]
\State Initialize $S_{2N-2}=0$
\For {$k = 2N-1, 2N, 2N+1, \ldots$}  
\State Gather the sample $(x_{k+1},u_{k})$
\State Compute $\hat{\Theta}^{ref}_{k}$ and $\hat{\Theta}^{test}_{k}$ as in \eqref{ref} and \eqref{test}, respectively
\State Compute $\gamma_{k}$  as in \eqref{gamma}
\If{$\|\hat{\Theta}^{ref}_{k}-\hat{\Theta}^{test}_{k}\|\geq \gamma_{k}$ and $k-S_{k-1}> 2N-2$} 
\State Set $S_{k}=k$
\State Flag $k$ as a change point
\Else
\State Set $S_{k}=S_{k-1}$
\EndIf
\EndFor
\end{algorithmic}
\end{algorithm}
\begin{remark}
One can update the least squares solution using the Sherman-Morrison formula \cite{henderson1981deriving}, which provides an efficient way to update the matrix inverse if the changes are `small' (in our case, the change at each time step is the replacement of the oldest sample by the most recent one). The threshold $\gamma_k$ will be provided in the next section, and depends on the parameters $b_{\sigma_{w}}$ and $b_{\Theta}$ (which depend on prior knowledge of the system, or can be estimated in practice). If those parameters are not available, one can replace $\gamma_{k}$ with any other positive value, but at the cost of losing performance guarantees. In general, a smaller threshold would lead to a higher probability of both false and true alarms. 
\end{remark}

In the next section, we will present our main results showing that Algorithm \ref{alg} ensures the false alarm probability will be less than $\delta$. We further provide a finite-sample-based lower bound on the true alarm probability, which demonstrates the trade-off between detection accuracy and detection delay.

\section{Finite Sample Analysis of Algorithm \ref{alg}}
In this section, we present theoretical guarantees for Algorithm \ref{alg}. Some of the proofs are included in the appendix. In Section \ref{sec:inter}, we present some intermediate results that are used later. Our main results (the finite-sample bounds on the probabilities of making false and true alarms) are presented in Section \ref{sec:main}. We first make the following assumption on system \eqref{system}.

\begin{assumption} \label{boundedV}
There exists a constant $\beta>0$ such that $\tr(\mathbf{E}[x_{k}x_{k}'])\leq \beta$ for all $k\geq 0$. 
\end{assumption}

If we have $\rho(A_{k})<1$ for all $k\geq 0$, i.e., all potential systems are strictly stable, Assumption \ref{boundedV} simply requires the dynamics to not change too frequently \cite{zhai2001stability} (as a sufficient condition) or that there are finitely many change points.

\subsection{Intermediate results}\label{sec:inter}
The lemma below provides a value for the threshold  $\gamma_{k}$, along with an associated probability bound on the recovered system matrices $\hat{\Theta}_k^{ref}$ and $\hat{\Theta}_k^{test}$.
\begin{lemma} \label{sysid}
 Consider any time step $k^*\geq 2N-1$. Let $\bar{V}_{k^*}^{ref}=(Z_{k^*}^{ref}(Z_{k^*}^{ref})'+V)V^{-1}$ and $\bar{V}_{k^*}^{test}=(Z_{k^*}^{test}(Z_{k^*}^{test})'+V)V^{-1}$, where $V=\lambda I_{n+p}$. Suppose
\begin{equation} 
\begin{aligned}
X^{ref}_{k^*}=\Theta_{k^*-2N+2} Z^{ref}_{k^*}+W^{ref}_{k^*},
\end{aligned}
\end{equation}
and 
\begin{equation} 
\begin{aligned}
X^{test}_{k^*}=\Theta_{k^*-N+2} Z^{test}_{k^*}+W^{test}_{k^*},
\end{aligned}
\end{equation}
i.e., the system matrices at the start of the reference interval and the test interval are the same throughout the interval, and that the threshold is chosen as 
\begin{equation}  \label{gamma}
\begin{aligned}
\gamma_{k^*}&= \frac{b_{\sigma_{w}}\sqrt{\frac{32}{9}(\log\frac{9^{n}2}{\delta}+\frac{1}{2}\log\det(\bar{V}_{k^*}^{ref}))}}{\sqrt{\lambda_{min}(Z_{k^*}^{ref}(Z_{k^*}^{ref})'+\lambda I_{n+p})}}\\
&+\frac{\lambda b_{\Theta}}{\lambda_{min}(Z_{k^*}^{ref}(Z_{k^*}^{ref})'+\lambda I_{n+p})}\\
&+ \frac{b_{\sigma_{w}}\sqrt{\frac{32}{9}(\log\frac{9^{n}2}{\delta}+\frac{1}{2}\log\det(\bar{V}_{k^*}^{test}))}}{\sqrt{\lambda_{min}(Z_{k^*}^{test}(Z_{k^*}^{test})'+\lambda I_{n+p})}}\\
&+\frac{\lambda b_{\Theta}}{\lambda_{min}(Z^{test}_{k^*}(Z^{test}_{k^*})'+\lambda I_{n+p})}.
\end{aligned}
\end{equation}
Then we have
\begin{equation*}
\begin{aligned}
&P(\|\hat{\Theta}^{ref}_{k^*}-\Theta_{k^*-2N+2}\|+\|\hat{\Theta}^{test}_{k^*}-\Theta_{k^*-N+2}\|\geq \gamma_{k^*})\\
&\quad \leq \delta.
\end{aligned}
\end{equation*}
\end{lemma}

\begin{pf}
We will focus on bounding the term $\|\hat{\Theta}^{ref}_{k^*}-\Theta_{k^*-2N+2}\|$ as the analysis for the term $\|\hat{\Theta}^{test}_{k^*}-\Theta_{k^*-N+2}\|$ is almost the same. Recalling the expression of $\hat{\Theta}^{ref}_{k^*}$ in \eqref{ref}, we have
\begin{equation}\label{touse4}
\begin{aligned}
&\|\hat{\Theta}^{ref}_{k^*}-\Theta_{k^*-2N+2}\|\\
&= \|W^{ref}_{k^*}(Z^{ref}_{k^*})'(Z^{ref}_{k^*}(Z^{ref}_{k^*})'+\lambda I_{n+p})^{-1}\\
&-\lambda\Theta_{k^*-2N+2}(Z^{ref}_{k^*}(Z^{ref}_{k^*})'+\lambda I_{n+p})^{-1}\|\\
&\leq \|W^{ref}_{k^*}(Z^{ref}_{k^*})'(Z^{ref}_{k^*}(Z^{ref}_{k^*})'+\lambda I_{n+p})^{-1}\|\\
&+\|\lambda\Theta_{k^*-2N+2}(Z^{ref}_{k^*}(Z^{ref}_{k^*})'+\lambda I_{n+p})^{-1}\|\\
&\leq \|W^{ref}_{k^*}(Z^{ref}_{k^*})'(Z^{ref}_{k^*}(Z^{ref}_{k^*})'+\lambda I_{n+p})^{-1}\|\\
&+\frac{\lambda b_{\Theta}}{\lambda_{min}(Z_{k^*}^{ref}(Z_{k^*}^{ref})'+\lambda I_{n+p})}.
\end{aligned}
\end{equation}
For the first term after the last inequality of \eqref{touse4}, we have
\begin{equation} \label{touse5}
\begin{aligned}
&\|W^{ref}_{k^*}(Z^{ref}_{k^*})'(Z^{ref}_{k^*}(Z^{ref}_{k^*})'+\lambda I_{n+p})^{-1}\|\\
&\leq \|W^{ref}_{k^*}(Z^{ref}_{k^*})'(Z^{ref}_{k^*}(Z^{ref}_{k^*})'+\lambda I_{n+p})^{-1/2}\| \times \\
&\qquad\qquad\qquad \qquad\qquad\|(Z^{ref}_{k^*}(Z^{ref}_{k^*})'+\lambda I_{n+p})^{-1/2}\|\\ 
&= \frac{\|W^{ref}_{k^*}(Z^{ref}_{k^*})'(Z^{ref}_{k^*}(Z^{ref}_{k^*})'+\lambda I_{n+p})^{-1/2}\|}{\sqrt{\lambda_{min}(Z^{ref}_{k^*}(Z^{ref}_{k^*})'+\lambda I_{n+p})}}\\
&=\frac{\|(\lambda I_{n+p}+\sum_{t=k^*-2N+2}^{k^*-N}z_{t}z_{t}')^{-1/2}(\sum_{t=k^*-2N+2}^{k^*-N} z_{t}w_{t}')\|}{\sqrt{\lambda_{min}(Z^{ref}_{k}(Z^{ref}_{k})'+\lambda I_{n+p})}}.
\end{aligned}
\end{equation}
Now we will bound the numerator of the last equality of \eqref{touse5} using Lemma \ref{martingale_bound_multi} in the appendix. Define the sequence pairs $\{\bar{z}_{t}\}_{t\geq1}$ and $\{\bar{w}_{t}\}_{t\geq1}$, where $\bar{z}_{t}=z_{k^*-2N+1+t}$ and $\bar{w}_{t}=w_{k^*-2N+1+t}$. Then we have $\|(\lambda I_{n+p}+\sum_{t=k^*-2N+2}^{k^*-N}z_{t}z_{t}')^{-1/2}(\sum_{t=k^*-2N+2}^{k^*-N} z_{t}w_{t}')\|=\|(\lambda I_{n+p}+\sum_{t=1}^{N-1}\bar{z}_{t}\bar{z}_{t}')^{-1/2}(\sum_{t=1}^{N-1} \bar{z}_{t}\bar{w}_{t}')\|$. Further, define the 
filtration $\{\mathcal{F}_{t}\}_{t\geq 0}$, where $\mathcal{F}_{t}=\sigma(\{\bar{z}_{i}\}_{i=1}^{1+t}\cup \{\bar{w}_{j}\}_{j=1}^{t})$. With these definitions, we have the noise terms $\bar{w}_{t}$ are $\mathcal{F}_{t}$-measurable, and $\bar{w}_{t}|\mathcal{F}_{t-1}$ are sub-Gaussian with parameter $\sigma_{w}^2$ for all $t\geq1$. Consequently,  fixing $\delta>0$, we can apply Lemma \ref{martingale_bound_multi} to obtain with probability at least $1-\frac{\delta}{2}$
\begin{equation}
\begin{aligned}
&\frac{\|(\lambda I_{n+p}+\sum_{t=k^*-2N+2}^{k^*-N}z_{t}z_{t}')^{-1/2}(\sum_{t=k^*-2N+2}^{k^*-N} z_{t}w_{t}')\|}{\sqrt{\lambda_{min}(Z^{ref}_{k^*}(Z^{ref}_{k^*})'+\lambda I_{n+p})}}\\
&\leq\frac{\sqrt{\frac{32}{9}\sigma_{w}^{2}(\log\frac{9^n 2}{\delta}+\frac{1}{2}\log\det(\bar{V}_{k^*}^{ref})}}{\sqrt{\lambda_{min}(Z^{ref}_{k^*}(Z^{ref}_{k^*})'+\lambda I_{n+p})}}.
\end{aligned}
\end{equation}
Combining the above inequality with \eqref{touse4}, using $\sigma_{w} \leq b_{\sigma_{w}}$, we have with probability at least $1-\frac{\delta}{2}$
\begin{equation}
\begin{aligned}
&\|\hat{\Theta}^{ref}_{k^*}-\Theta_{k^*-2N+2}\|\\
&\leq \frac{b_{\sigma_{w}}\sqrt{\frac{32}{9}(\log\frac{9^{n}2}{\delta}+\frac{1}{2}\log\det(\bar{V}_{k^*}^{ref}))}}{\sqrt{\lambda_{min}(Z_{k^*}^{ref}(Z_{k^*}^{ref})'+\lambda I_{n+p})}}\\
&+\frac{\lambda b_{\Theta}}{\lambda_{min}(Z_{k^*}^{ref}(Z_{k^*}^{ref})'+\lambda I_{n+p})}.
\end{aligned}
\end{equation}
Following a similar procedure for the term $\|\hat{\Theta}^{ref}_{k^*}-\Theta_{k^*-N+2}\|$, and applying a union bound, we have the desired result.
\end{pf}

The following lemma bounds the probability of the metric $\|\hat{\Theta}^{ref}_{k}-\hat{\Theta}^{test}_{k}\|$ being larger than the threshold $\gamma_{k}$ if there is not a change point in both the reference interval and the test interval. 

\begin{lemma} \label{lemma:false alarm}
Consider any time step $k^*\geq 2N-1$. If it holds that $\Theta_{k^*}=\Theta_{k^*-1}=\ldots =\Theta_{k^*-2N+1}$, 
then we have
\begin{equation*}
\begin{aligned}
P(\|\hat{\Theta}^{ref}_{k^*}-\hat{\Theta}^{test}_{k^*}\|\geq \gamma_{k^*})\leq \delta.
\end{aligned}
\end{equation*}
\end{lemma}

\begin{pf}
Since $\Theta_{k^*}=\Theta_{k^*-1}=\ldots =\Theta_{k^*-2N+1}$, we have
\begin{equation*} 
\begin{aligned}
&\|\hat{\Theta}^{ref}_{k^*}-\hat{\Theta}^{test}_{k^*}\|\\
&=\|(\hat{\Theta}^{ref}_{k^*}-\Theta_{k^*})-(\hat{\Theta}^{test}_{k^*}-\Theta_{k^*})\|\\
&\leq \|\hat{\Theta}^{ref}_{k^*}-\Theta_{k^*}\|+\|\hat{\Theta}^{test}_{k^*}-\Theta_{k^*}\|\\
&= \|\hat{\Theta}^{ref}_{k^*}-\Theta_{k^*-2N+2}\|+\|\hat{\Theta}^{test}_{k^*}-\Theta_{k^*-N+2}\|.
\end{aligned}
\end{equation*}
Further, note that we also have
\begin{equation*}
\begin{aligned}
&X^{ref}_{k^*}=\Theta_{k^*-2N+2} Z^{ref}_{k^*}+W^{ref}_{k^*},
\end{aligned}
\end{equation*}
\begin{equation*}
\begin{aligned}
&X^{test}_{k^*}=\Theta_{k^*-N+2} Z^{test}_{k^*}+W^{test}_{k^*}.
\end{aligned}
\end{equation*}
Applying Lemma \ref{sysid}, we get the desired result.
\end{pf}

The following two lemmas are used to establish our finite-sample bound on the true alarm probability, with proofs provided in the appendix.

\begin{lemma}\label{lower}
Let $k^*$ be a time step such that $k^*\geq 2N-1$. For any fixed $\bar{\delta}>0$,  let $N\geq \max(42,200(n+p) \log(\frac{13}{\bar{\delta}}))$. Then with probability at least $1-2\bar{\delta}$, the following inequalities hold simultaneously:
\begin{equation*}
Z^{ref}_{k^*}(Z^{ref}_{k^*})'+\lambda I_{n+p}\succeq \frac{N\sigma_{min}^2+42\lambda}{42} I_{n+p},
\end{equation*}
\begin{equation*}
Z^{test}_{k^*}(Z^{test}_{k^*})'+\lambda I_{n+p}\succeq \frac{N\sigma_{min}^2+42\lambda}{42} I_{n+p}.
\end{equation*}
\end{lemma}

\begin{lemma} \label{markov}
Let $k^*$ be a time step such that $k^*\geq 2N-1$. For any fixed $\bar{\delta}>0$, with probability at least $1-2\bar{\delta}$, the following inequalities hold simultaneously:
\begin{equation*}
\begin{aligned}
&\|Z_{k^*+N-2}^{ref}(Z_{k^*+N-2}^{ref})'\|\leq \frac{C_{1}}{\bar{\delta}},
\end{aligned} 
\end{equation*}
\begin{equation*}
\begin{aligned}
&\|Z_{k^*+N-2}^{test}(Z_{k^*+N-2}^{test})'\|\leq \frac{C_{1}}{\bar{\delta}},
\end{aligned}
\end{equation*}
where $C_{1}=(N-1)(\beta+\sigma_{u}^2 p)$.
\end{lemma}

The following result bounds the probability of the threshold value $\gamma_{k}$ being small compared to the magnitude of the change, and will be used later to lower bound the probability of making an accurate detection.

\begin{lemma} \label{deltae}
Let $k^*\geq 2N-1$ be a change point. Let $N_{1}=200(n+p)(\log(\frac{7\lambda}{C_{1}})+\|\Delta_{k^*}\|\sqrt{\frac{N\sigma_{min}^2+42\lambda}{2500b_{\sigma_{w}}^2(n+p)}}-\frac{168\lambda b_{\Theta}}{\sqrt{2500b_{\sigma_{w}^2}(n+p)(N\sigma_{min}^2+42\lambda)}})$. 
Consider any $N$ satisfying $N\geq \max(42,\allowbreak N_{1},\frac{336\lambda b_{\Theta}}{\|\Delta_{k^{*}}\|\sigma_{min}^2}-\frac{42\lambda}{\sigma_{min}^2})$. Further let $\lambda\leq \frac{4C_{1}}{e\delta_{e}}$, where 
\begin{equation*}
\begin{aligned} 
\delta_{e}=\frac{8C_{1}}{\lambda \exp(\|\Delta_{k^*}\|\sqrt{\frac{N\sigma_{min}^2+42\lambda}{10000b_{\sigma_{w}}^2(n+p)}}-\sqrt{\frac{\log(\frac{9^n 2}{\delta})}{n+p}})}.
\end{aligned}
\end{equation*}
Then we have
\begin{equation*}
\begin{aligned}
P(\{2\gamma_{k^*+N-2}\leq \|\Delta_{k^*}\|\}) \geq 1-\delta_{e},
\end{aligned}
\end{equation*}
where $\gamma_{k^*+N-2}$ is defined in Lemma \ref{sysid}, and $C_{1}$ is defined in Lemma \ref{markov}.
\end{lemma}

\begin{pf}
Fix $\bar{\delta}>0$. From Lemma \ref{lower}, when $N\geq \max(42,200(n+p) \log(\frac{13}{\bar{\delta}}))$, we have with probability at least $1-2\bar{\delta}$ the following inequalities
\begin{equation} \label{event1}
\begin{aligned}
\lambda_{min}(Z_{k^*+N-2}^{ref}(Z_{k^*+N-2}^{ref})'+\lambda I_{n+p})\geq \frac{N\sigma_{min}^2+42\lambda}{42},
\end{aligned}
\end{equation}
\begin{equation} \label{event2}
\begin{aligned}
\lambda_{min}(Z_{k^*+N-2}^{test}(Z_{k^*+N-2}^{test})'+\lambda I_{n+p})\geq \frac{N\sigma_{min}^2+42\lambda}{42}.
\end{aligned}
\end{equation}
From Lemma \ref{markov}, letting $\lambda \leq \frac{C_{1}}{e\bar{\delta}}$, we have with probability at least $1-2\bar{\delta}$
\begin{equation} \label{event3}
\begin{aligned}
\det(\bar{V}_{k^*+N-2}^{ref})&= \frac{\det(Z_{k^*+N-2}^{ref}(Z_{k^*+N-2}^{ref})'+V)}{\det(V)}\\
& \leq \frac{(\|Z_{k^*+N-2}^{ref}(Z_{k^*+N-2}^{ref})'\|+\lambda)^{n+p}}{\lambda^{n+p}}\\
&\leq (\frac{C_{1}}{\bar{\delta}\lambda}+1)^{n+p}\leq (\frac{2C_{1}}{\bar{\delta}\lambda})^{n+p},
\end{aligned}
\end{equation}
and
\begin{equation} \label{event4}
\begin{aligned}
\det(\bar{V}_{k^*+N-2}^{test})\leq (\frac{2C_{1}}{\bar{\delta}\lambda})^{n+p},
\end{aligned}
\end{equation}
where $\bar{V}_{k^*+N-2}^{ref}$ and $\bar{V}_{k^*+N-2}^{test}$ are defined in Lemma \ref{sysid}. Applying a union bound, we have the events in \eqref{event1}-\eqref{event4} occur simultaneously with probability at least $1-4\bar{\delta}$, which implies
\begin{equation*}  
\begin{aligned}
2\gamma_{k^*+N-2}&\leq \frac{50b_{\sigma_{w}}\sqrt{\log(\frac{9^n 2}{\delta})}+50b_{\sigma_{w}}\log(\frac{2C_{1}}{\bar{\delta}\lambda})\sqrt{n+p}}{\sqrt{N\sigma_{min}^2+42\lambda}}\\
&+\frac{168\lambda b_{\Theta}}{N\sigma_{min}^2+42\lambda},
\end{aligned}
\end{equation*}
where we used the relationship that $\sqrt{a+b}\leq \sqrt{a}+\sqrt{b}$ for positive $a,b$, and that $\lambda \leq \frac{C_{1}}{e\bar{\delta}}\Rightarrow \frac{2C_{1}}{\bar{\delta}\lambda}\geq e\Rightarrow \sqrt{\log(\frac{2C_{1}}{\bar{\delta}\lambda})}\leq \log(\frac{2C_{1}}{\bar{\delta}\lambda})$.

Setting the right hand side of the above inequality to $\|\Delta_{k^*}\|$, after some algebraic manipulations, we have
\begin{equation}  \label{touse8}
\begin{aligned}
\bar{\delta}=\frac{2C_{1}}{\lambda\exp(\frac{c}{50b_{\sigma_{w}}\sqrt{n+p}})},
\end{aligned}
\end{equation}
where $c=\|\Delta_{k^*}\|\sqrt{N\sigma_{min}^2+42\lambda}-\frac{168\lambda b_{\Theta}}{\sqrt{N\sigma_{min}^2+42\lambda}}-50b_{\sigma_{w}}\sqrt{\log(\frac{9^n 2}{\delta})}$. When $N\geq \frac{336\lambda b_{\Theta}}{\|\Delta_{k^{*}}\|\sigma_{min}^2}-\frac{42\lambda}{\sigma_{min}^2}$, we have $c\geq \frac{\|\Delta_{k^*}\|\sqrt{N\sigma_{min}^2+42\lambda}}{2}-50b_{\sigma_{w}}\sqrt{\log(\frac{9^n 2}{\delta})}$, which implies $\bar{\delta}\leq \frac{\delta_{e}}{4}$. Finally, recall that we also required the conditions $N\geq \max(42,200(n+p) \log(\frac{13}{\bar{\delta}}))$ and $\lambda \leq \frac{C_{1}}{e\bar{\delta}}$. Substituting \eqref{touse8} into these conditions, using $\bar{\delta}\leq \frac{\delta_{e}}{4}$, and after some simplifications, we have the desired result. 
\end{pf}

\begin{remark} \label{expdeltae}
In Theorem \ref{thm:true detection}, we will show that a smaller $\delta_{e}$ corresponds to a higher probability of true alarm. Note that $N$ will be larger than $N_{1}$ for sufficiently large $N$ since the term  $C_{1}$ grows linearly fast. 
\end{remark}

The following result lower bounds the probability of the metric $\|\hat{\Theta}^{ref}_{k}-\hat{\Theta}^{test}_{k}\|$ being larger than the threshold $\gamma_{k}$ with some delay, when there is a change point.
\begin{lemma}\label{lemma:true alarm}
Consider any time step $k^*$ that is a sufficiently separated change point. Suppose the conditions in Lemma \ref{deltae} are satisfied.  
Then we have
\begin{equation*}
\begin{aligned}
P(\bigcup_{t=k^*}^{k^*+N-2}\{\|\hat{\Theta}^{ref}_{t}-\hat{\Theta}^{test}_{t}\|\geq \gamma_{t}\}) \geq 1-\delta-\delta_{e},
\end{aligned}
\end{equation*}
where $\delta_{e}$ is defined in Lemma \ref{deltae}.
\end{lemma}

\begin{pf}
Note that we have 
\begin{equation*}
\begin{aligned}
&P(\bigcup_{t=k^*}^{k^*+N-2}\{\|\hat{\Theta}^{ref}_{t}-\hat{\Theta}^{test}_{t}\|\geq \gamma_{t}\})\\
&\geq P(\{\|\hat{\Theta}^{ref}_{k^*+N-2}-\hat{\Theta}^{test}_{k^*+N-2}\|\geq \gamma_{k^*+N-2}\}) .
\end{aligned}
\end{equation*}

We will lower bound the probability after the above inequality now. We have
\begin{equation} \label{touse6}
\begin{aligned}
&\|\hat{\Theta}^{ref}_{k^*+N-2}-\hat{\Theta}^{test}_{k^*+N-2}\|\\
&=\|(\Theta_{k^*-1}+\hat{\Theta}^{ref}_{k^*+N-2}-\Theta_{k^*-1})-\\
&\quad\quad \quad  \quad \quad \quad (\Theta_{k^*}+\hat{\Theta}^{test}_{k^*+N-2}-\Theta_{k^*})\|\\
&=\|\Delta_{k^*}+(\hat{\Theta}^{ref}_{k^*+N-2}-\Theta_{k^*-1})-(\hat{\Theta}^{test}_{k^*+N-2}-\Theta_{k^*})\|\\
&\geq \|\Delta_{k^*}\|-\|\hat{\Theta}^{ref}_{k^*+N-2}-\Theta_{k^*-1}\|-\|\hat{\Theta}^{test}_{k^*+N-2}-\Theta_{k^*}\|,\\
\end{aligned}
\end{equation}
where the last inequality is due to the triangle inequality. Since $k^*$ is a sufficiently separated change point, we have $\Theta_{k^*-1}=\Theta_{k^*-2}=\ldots=\Theta_{k^*-4N+1}$ and $\Theta_{k^*}=\Theta_{k^*+1}=\ldots=\Theta_{k^*+4N-2}$. Hence, we have
\begin{equation}  \label{touse1}
\begin{aligned}
&\|\hat{\Theta}^{ref}_{k^*+N-2}-\Theta_{k^*-1}\|=\|\hat{\Theta}^{ref}_{k^*+N-2}-\Theta_{k^*-N}\|.
\end{aligned}
\end{equation}
For the same reason, we have
\begin{equation} 
\begin{aligned}
X^{ref}_{k^*+N-2}=\Theta_{k^*-N} Z^{ref}_{k^*+N-2}+W^{ref}_{k^*+N-2},
\end{aligned}
\end{equation}
and 
\begin{equation} 
\begin{aligned}
X^{test}_{k^*+N-2}=\Theta_{k^*} Z^{test}_{k^*+N-2}+W^{test}_{k^*+N-2}.
\end{aligned}
\end{equation}
Applying Lemma \ref{sysid}, we can obtain that with probability at least $1-\delta$
\begin{equation} 
\begin{aligned}
\|\hat{\Theta}^{ref}_{k^*+N-2}-\Theta_{k^*-N}\|+\|\hat{\Theta}^{test}_{k^*+N-2}-\Theta_{k^*}\|\leq \gamma_{k^*+N-2},
\end{aligned}
\end{equation}
which implies that 
\begin{equation} 
\begin{aligned}
\|\hat{\Theta}^{ref}_{k^*+N-2}-\hat{\Theta}^{test}_{k^*+N-2}\|\geq \|\Delta_{k^*}\|-\gamma_{k^*+N-2}
\end{aligned}
\end{equation}
with probability at least $1-\delta$ from \eqref{touse6} and \eqref{touse1}. Furthermore, applying Lemma \ref{deltae} and a union bound, we have with probability at least $1-\delta-\delta_{e}$
\begin{equation*} 
\begin{aligned}
\|\hat{\Theta}^{ref}_{k^*+N-2}-\hat{\Theta}^{test}_{k^*+N-2}\|&\geq \|\Delta_{k^*}\|-\gamma_{k^*+N-2}\\
&\geq \gamma_{k^*+N-2}.
\end{aligned}
\end{equation*}
\end{pf}

\subsection{Main results: Finite-sample probability bounds on making false and true alarms} \label{sec:main}
Now, we state our first main theorem, which shows that the the probability of false alarm is upper bounded by $\delta$ (recall that setting $S_{k}=k$ is always associated with flagging time step $k$ as a change point).

\begin{thm}[Probability of False Alarm] \label{thm:false detection}
Consider any time step $k^*\geq 2N-1$. If it holds that $\Theta_{k^*}=\Theta_{k^*-1}=\ldots =\Theta_{k^*-2N+1}$, we have
\begin{equation*}
\begin{aligned}
P(S_{k^*}=k^*) \leq \delta.
\end{aligned}
\end{equation*}
\end{thm}

\begin{pf}
From Algorithm \ref{alg}, we have $P(S_{k^*}=k^*)\leq P(\|\hat{\Theta}^{ref}_{k^*}-\hat{\Theta}^{test}_{k^*}\|\geq \gamma_{k^*})$. Applying Lemma \ref{lemma:false alarm}, we have $P(S_{k^*}=k^*)\leq \delta$. 
\end{pf}

\begin{remark}
Note that $\delta$ is a user specified parameter that can be arbitrarily small. However, a small $\delta$ could reduce the true alarm probability, which will be discussed when we present Theorem \ref{thm:true detection}. We also note that although we assumed that both the input $u_{k}$ and the noise $w_{k}$ are i.i.d Gaussian, this result actually holds for independent sub-Gaussian noise and arbitrary input that is independent of future noise. However, our finite-sample lower bound of the true alarm probability, i.e., if there is a change point, requires the assumption of i.i.d Gaussian input and i.i.d Gaussian noise. We leave the analysis of more general distributions of input and noise to future work.
\end{remark}

Next, we state our result that lower bounds the probability of making a true alarm (with some delay).
\begin{thm}[Probability of True Alarm] \label{thm:true detection}
Consider any time step $k^*$ that is a sufficiently separated change point. 
Suppose the conditions on $N,\lambda$ in Lemma \ref{deltae} are also satisfied. 
Recall that $C_{1}=(N-1)(\beta+\sigma_{u}^2 p)$, and
\begin{equation*}
\begin{aligned} 
\delta_{e}=\frac{8C_{1}}{\lambda \exp(\|\Delta_{k^*}\|\sqrt{\frac{N\sigma_{min}^2+42\lambda}{10000b_{\sigma_{w}}^2(n+p)}}-\sqrt{\frac{\log(\frac{9^n 2}{\delta})}{n+p}})}.
\end{aligned}
\end{equation*}
Then we have
\begin{equation*}
\begin{aligned}
&P(\bigcup_{t=k^*}^{k^*+N-2}\{S_{t}=t\})\geq 1- (2N+1)\delta-\delta_{e},
\end{aligned}
\end{equation*}
where the term $(2N+1)\delta$ captures the uncertainty due to settling from the previous change point, and the term $\delta_{e}$ captures the uncertainty due to noise.
\end{thm}


\begin{pf}
Define the events $E_{1}\triangleq\bigcup_{t=k^*}^{k^*+N-2}\{\|\hat{\Theta}^{ref}_{t}-\hat{\Theta}^{test}_{t}\|\geq \gamma_{t}\}$ and $E_{2}\triangleq\bigcap_{t=k^*-2N}^{k^*-1}\{\|\hat{\Theta}^{ref}_{t}-\hat{\Theta}^{test}_{t}\|\leq \gamma_{t} \}$. We have $P(E_{1})\geq 1-\delta-\delta_{e}$ by applying Lemma \ref{lemma:true alarm}. Since $k^*$ is a sufficiently separated change point, we also have $\Theta_{k^*-1}=\Theta_{k^*-2}=\ldots=\Theta_{k^*-4N+1}$. Hence we can apply Lemma \ref{lemma:false alarm} and combine the $2N$ events in $E_{2}$ using a union bound to obtain $P(E_{2})\geq 1-2N\delta$. Recall the conditions for setting $S_{t}=t$ in Algorithm \ref{alg}. Conditioning on $E_{1}\cap E_{2}$, let $\hat{t}$ be the smallest time step such that $\|\hat{\Theta}^{ref}_{t}-\hat{\Theta}^{test}_{t}\|\geq \gamma_{t}$ for $t=k^*,\ldots, k^*+N-2$. Note that on the event $E_{1} \cap E_{2}$ we also have
\begin{equation*}
\begin{aligned}
\hat{t}-S_{\hat{t}-1}&\geq k^*-S_{k^*-2N-1}\geq k^*-(k^*-2N-1)\\
&> 2N-2,
\end{aligned}
\end{equation*}
with probability $1$, which implies $S_{\hat{t}}=\hat{t}$.
Consequently, combining events $E_{1}$ and $E_{2}$ using a union bound, we have
\begin{equation*}
\begin{aligned}
P(\bigcup_{t=k^*}^{k^*+N-2}\{S_{t}=t\})&\geq P(E_{1}\cap E_{2} )\\
&\geq 1-(2N+1)\delta-\delta_{e}.
\end{aligned}
\end{equation*}
\end{pf}

\begin{remark} \label{discussions}
\textbf{Interpretation of Theorem \ref{thm:true detection}}. Recall that setting $S_{t}=t$ implies flagging time step $t$ as a change point. Theorem \ref{thm:true detection} provides a high probability lower bound on true detection in the long run (i.e., predict the change point after $k^*$, but within any pre-specified large 
time steps), where the associated delay is bounded by $N-2$. Note that the term $-2N\delta$ is added to the probability lower bound in Lemma \ref{lemma:true alarm}. This term is included due to the need to deal with potentially multiple change points, since the algorithm may conclude that the dynamics have settled by observing the event $E_{2}=\bigcap_{t=k^*-2N}^{k^*-1}\{\|\hat{\Theta}^{ref}_{t}-\hat{\Theta}^{test}_{t}\|\leq \gamma_{t} \}$, i.e., we want to make sure that the reason the metric is greater than the threshold is that we have a new change point instead of that we are observing some residual effects from the previous change point. Below we further discuss the effects of some parameters.

\textbf{Discussions on the effects of $\delta, N$:} Suppose that the window size $N$ is fixed for now. We see from Theorem \ref{thm:true detection} that a smaller false probability $\delta$ would lead to a smaller uncertainty due to the need to confirm the dynamics has settled from the previous change point. However, if $\delta$ is set to be too small, $\delta_{e}$ may become very large and make the overall lower bound small, which implies a potentially smaller probability of true detection in the long run. 
To make the true alarm probability lower bound large and maintain a small false alarm probability $\delta$, we can use a larger window size $N$. More specifically, 
by setting $\delta=\frac{a}{\exp(\sqrt{N})}$ for some constant $a>0$, we see that a larger $N$ corresponds to a smaller false alarm probability, and ensures both a smaller uncertainty due to settling and a smaller uncertainty due to noise, since the term $C_{1}$ in the numerator of $\delta_{e}$ grows at most linearly with respect to $N$, and the denominator of $\delta_{e}$ grows exponentially fast with respect to the square root of $N$. The price, on the other hand, is that the guaranteed delay is larger due to a larger $N$. Consequently, this result demonstrates a trade-off between detection accuracy and detection delay, i.e., to maintain both a low false alarm probability and a high true alarm probability in the long run, one has to suffer from a potentially larger delay by using a larger $N$. Such a non-asymptotic characterization is different from existing approaches, where algorithm performance is typically measured in expectation, e.g., using Average Running Length \cite{alanqary2021change}.  
Furthermore, as $N$ becomes larger, there may be fewer change points that satisfy the sufficient separation condition, i.e., the speed of the change of the dynamics may appear to be too fast relative to the time interval we are monitoring. In other words, this result further implies that a smaller $N$ is suitable for more frequent changes, but at the price of being less likely to predict the change accurately.  

\textbf{Discussions on the effects of $\|\Delta_{k^*}\|$:} Finally, note that the denominator of $\delta_{e}$ also grows exponentially fast with respect to the magnitude of change $\|\Delta_{k^*}\|$ (supposing that $N$ is large enough). Hence, a larger change in dynamics could lead to a higher probability (lower bound) of detecting a change point (matching intuition).

\end{remark}

Theorem \ref{thm:false detection} and Theorem \ref{thm:true detection} cover all possible situations if all change points are sufficiently separated. More specifically, consider any time step $k^*\geq 2N-1$. If there is not a change point over the past $2N$ time steps, Theorem \ref{thm:false detection} ensures that there will not be a false alarm with high probability. If the current time step $k^*$ is a change point, Theorem \ref{thm:true detection} ensures that Algorithm \ref{alg} will detect it, i.e., predict exactly one change point, within $N-1$ time steps, with high probability. Furthermore, the design of Algorithm \ref{alg} ensures that the Algorithm will not flag any change point at $k^*+N-1, k^*+N,\ldots, k^*+2N-2$, if a change point was predicted at $k^*,\ldots,k^*+N-2$ (note that the events $\{S_{t}=t\}_{t=k^*}^{k^*+2N-2}$ are mutually exclusive, since it is necessary to have $t-S_{t-1}>2N-2$ to set $S_{t}=t$ from Algorithm \ref{alg}). These properties imply that all time steps are covered by either Theorem \ref{thm:false detection} or Theorem \ref{thm:true detection}. 

\section{Numerical experiment}
\label{exp}
In this section, we provide some numerical examples of the online change point detection algorithm. In general, a threshold that has non-asymptotic guarantees can be conservative in practice. However, as we will see, even if we use the exact theoretical threshold, Algorithm \ref{alg} can still achieve reasonably good performance, which indicates that the derived threshold is not overly conservative in practice (and it does not require perfect knowledge of the parameters at any time, unlike \cite{lai1998information,geng2019online, banerjee2015data}).

The system we consider here is the linearized longitudinal dynamics of a UAV reported in \cite{ahmed2015modeling}, where we set the sampling rate to be 0.1 seconds using zero-order hold. The system has 5 states, representing inertial velocity components of the airframe projected onto a body frame axis, the pitch angle, the pitch angular rate, and the altitude. The input is the elevator deflection. Assuming full state observation, the  matrices $A_{k}$ and $B_{k}$ are given by
\begin{equation*}
\begin{aligned}
&A_{k}=\\
&\begin{bmatrix}
0.9371+\epsilon^{A}_{k} & 0.068&   -0.9507&   -0.0367&    0\\
 -0.0085&  0.2761&   -0.0207&    0.411&   0\\
0.0035  & -0.0164 &   0.9991&    0.043&    0\\
 0.0548 &  -0.1914&   -0.0253&    0.0593&    0\\
-0.0086 &   0.0726&   -1.6984&   -0.0146&    1\\
\end{bmatrix},
\end{aligned}
\end{equation*}
\begin{equation*}
\begin{aligned}
&B_{k}=\\
&\begin{bmatrix}
0.361+\epsilon^{B}_{k}&-4.8436&-0.3888&-5.6967&0.0492
\end{bmatrix}',
\end{aligned}
\end{equation*}
where $\epsilon^{A}_{k}$ and $\epsilon^{B}_{k}$ are perturbations we added to to the system, and we set $\epsilon^{A}_{k}=0,-1,-1$ for $0\leq k\leq 2499$, $2500\leq k\leq 4999$, and $k\geq 5000$, respectively; $\epsilon^{B}_{k}=0,2,0$ for $0\leq k\leq 2499$, $2500\leq k\leq 4999$, and $k\geq 5000$, respectively. In other words, $k_{1}=2500$ and $k_{2}=5000$ are two change points. We set $\lambda=\sigma_{w}=\sigma_{u}=1$, and  $x_{0}=0$. The parameters used in our threshold value $\gamma_{k}$ are assumed to be tight bounds for simplicity, i.e., $b_{\Theta}=\max_{k\geq 0}(\|\Theta_{k}\|)$ and $b_{\sigma_{w}}=\sigma_{w}$. We performed experiments using $N=50,150, 250,350, 450$, each with 10 independent runs, where the length of each experiment is set to $9000$ steps. The bound on false alarm probability is set to $\delta=\frac{1000}{\exp{\sqrt{N}}}$ (as suggested in Remark \ref{discussions}).

\begin{figure}
\minipage[t]{0.4\textwidth} 
    \includegraphics[width=\linewidth]{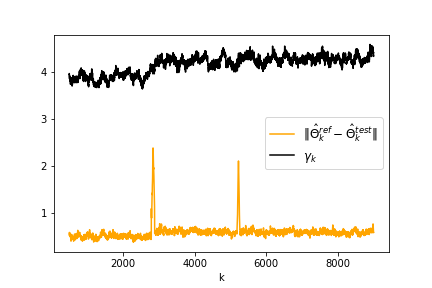}
    \caption{Online Change Point Detection with $N=50$. The use of small $N$ results in a threshold that is too high to flag change points, although we see spikes in the test statistics.}
    \label{N50} 
\endminipage \hfill
\minipage[t]{0.4\textwidth}
    \includegraphics[width=\linewidth]{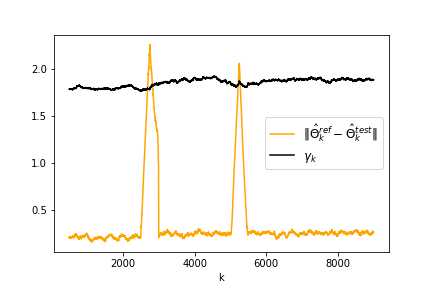}
    \caption{Online Change Point Detection with $N=250$. The threshold successfully captures the two change points using a moderate $N$.}
    \label{N250} 
\endminipage \hfill
\minipage[t]{0.4\textwidth}
    \includegraphics[width=\linewidth]{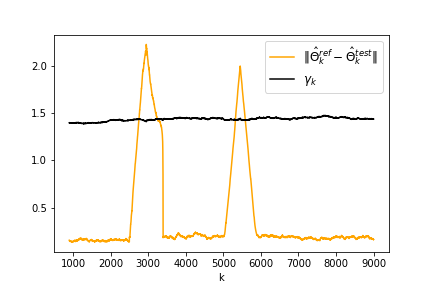}
    \caption{Online Change Point Detection with $N=450$. The threshold successfully captures the two change points, but the use of a larger $N$ incurs higher delay. }
    \label{N450} 
\endminipage
\end{figure}

In Figures \ref{N50}-\ref{N450}, we plot the average $\|\hat{\Theta}^{ref}_{k}-\hat{\Theta}^{test}_{k}\|$ versus $\gamma_{k}$ for $N=50,250,450$. In Table \ref{ADL}, we report the performance of the CPD algorithm. Here, AD1 refers to the average detection time between $k=2500$ and $k=4999$, AD2 refers to the average detection time between $k=5000$ and $k=8999$, MD1 refers to the number of experiments where no detection was made between $k=2500$ and $k=4999$, and MD2 refers to the number of experiments where no detection was made between $k=5000$ and $k=8999$. For all experiments, none of them made a false alarm before $k=2500$. Further, we can see that as $N$ increases, the probability of true detection in the long run increases (captured by a smaller misdetection rate). However, the corresponding delay also increases accordingly. These empirical observations are consistent with our theoretical findings in Theorem \ref{thm:true detection}. 

\begin{table}[ht!]
\centering
\begin{tabular}{||l c c c c r||} 
 \hline
$N$ &$\delta$ & AD1 & MD1 &AD2 & MD2\\ 
 \hline\hline
50  & 0.85 & 2550 &  9 & N/A & 10 \\
150 & $4.8\times 10^{-3}$ & 2629.3 &  1 & N/A & 10 \\
250 & $1.3\times 10^{-4}$  & 2685.8 &  0 & 5218.9 & 1 \\
350  &$7.5\times 10^{-6}$  & 2701.2&  0 & 5280.9 & 0 \\
450 & $6.1\times 10^{-7}$  & 2755&  0 & 5321.8 & 0 \\
 \hline
\end{tabular}
\caption{Empirical Performance of Algorithm \ref{alg} Over 10 Independent Runs. The bound on false alarm probability is set to $\delta=\frac{1000}{\exp{\sqrt{N}}}$.}
\label{ADL}
\end{table}

\section{Conclusion}
In this paper, we studied online change point detection for linear dynamical systems, where there are potentially multiple change points. Our analysis provides a data-dependent dynamic threshold that allows the user to specify a desired upper bound of the false alarm probability. We also provided a finite-sample lower bound on the probability of correctly identifying the change point with some delay.  Our analysis demonstrates the trade-off between detection accuracy and detection delay, and characterizes how frequently changes can occur while still maintaining detection with a given probability. It is noted that our focus in this paper is on fully-observed systems, i.e., all system states can be perfectly measured. It would be of interest to extend our analysis to partially-observed systems, where only a subset of system states can be measured. Other promising directions for future work would be to analyze different types of changes for dynamical systems, e.g., changes in noise distribution or changes in models that are possibly state-dependent, and to consider nonlinear system dynamics. 

\section*{Appendix}
\subsection{Auxiliary results}

The following lemma generalizes the self-normalized martingales in \cite{abbasi2011improved} to the multi-dimensional case. The proof can be found in \cite[Lemma~5]{xin2023learning}.

\begin{lemma} \label{martingale_bound_multi}
Let $\{\mathcal{F}_{t}\}_{t\geq 0}$ be a filtration. Let $\{{w}_{t}\}_{t\geq 1}$ be a  $\mathbb{R}^{n}$-valued stochastic process such that $w_{t}$ is $\mathcal{F}_{t}$-measurable, and $w_{t}$ is conditionally sub-Gaussian on $\mathcal{F}_{t-1}$ with parameter $R^2$. Let $\{z_{t}\}_{t\geq 1}$ be a $\mathbb{R}^{m}$-valued stochastic process such that $z_{t}$
is $\mathcal{F}_{t-1}$-measurable. Assume that $V$ is a $m\times m$ dimensional positive definite matrix. For all $t\geq 0$, define
\begin{equation*}
\begin{aligned}
&\bar{V}_{t}=V+\sum_{s=1}^{t}z_{s}z_{s}', S_{t}=\sum_{s=1}^{t}z_{s}w_{s}'.\\
\end{aligned}
\end{equation*}
Then, for any $\delta>0$, and for all $t\geq0$,
\begin{equation*}
\begin{aligned}
&P(\|\bar{V}_{t}^{-\frac{1}{2}}S_{t}\|\leq\sqrt{\frac{32}{9}R^{2}(\log\frac{9^n}{\delta}+\frac{1}{2}\log\det(\bar{V}_{t}V^{-1})})\\
&\geq 1-\delta.\\
\end{aligned}
\end{equation*}
\end{lemma}

The following result is used to establish Lemma \ref{lower}. 
\begin{lemma}\cite[Lemma~36]{cassel2020logarithmic}
Let $\{z_{t}\}_{t\geq 0}$ be a sequence of random vectors that is adapted to a filtration $\{\mathcal{F}_{t}\}_{t\geq 0}$, where $z_{t}\in \mathbb{R}^{d}$ for all $t\ge 0$. Suppose $z_{t}$ is conditionally Gaussian on $\mathcal{F}_{t-1}$ with $\mathbb{E}[z_{t}z_{t}'|\mathcal{F}_{t-1}]\succeq \sigma_{z}^2 I_{d}$ for all $t\ge 1$, where $\sigma_{z}>0$. Then, for any fixed $\delta>0$ and any $T\geq 200d \log(\frac{12}{\delta})$, the following inequality holds with probability at least $1-\delta$:
\begin{equation*}
\sum_{t=0}^{T-1}z_{t}z_{t}' \succeq \frac{(T-1)\sigma_{z}^2}{40}I_{d}.
\end{equation*}

\label{lemma:PE}
\end{lemma}

\subsection{Proof of Lemma \ref{lower}}
\begin{pf}
We will only show the first inequality as the proof for the second one is almost the same. Note that 
\begin{equation*}
Z^{ref}_{k^*}(Z^{ref}_{k^*})'=\sum_{t=k^*-2N+2}^{k^*-N}z_{t}z_{t}'.
\end{equation*}
Rename the sequences $\bar{z}_{t}=z_{k^*-2N+2+t}$, $\bar{u}_{t}=u_{k^*-2N+2+t}$, $\bar{w}_{t}=w_{k^*-2N+2+t}$, and $\bar{\Theta}_{t}=\Theta_{k^*-2N+2+t}$ for $t\geq 0$. Define the filtration $\{\mathcal{F}_{t}\}_{t\geq 0}$, where $\mathcal{F}_{t}=\sigma(\bar{z}_{0},\bar{z}_{1},\ldots,\bar{z}_{t})$ for $t\geq 0$. Note that $z_{t}|\mathcal{F}_{t-1}$ is a Gaussian random vector for $t\geq 1$, since 
\begin{equation*}
\begin{aligned}
\bar{z}_{t}|\mathcal{F}_{t-1}=
\begin{bmatrix}
\bar{\Theta}_{t-1} \bar{z}_{t-1}|\mathcal{F}_{t-1}\\
0\end{bmatrix}
+
\begin{bmatrix}
\bar{w}_{t-1}|\mathcal{F}_{t-1}\\
\bar{u}_{t}|\mathcal{F}_{t-1}\end{bmatrix}.
\end{aligned}
\end{equation*} 
From the above equality, we also have
\begin{equation*}
\begin{aligned}
\mathbb{E}[\bar{z}_{t}\bar{z}_{t}'|\mathcal{F}_{t-1}]
&\succeq 
\begin{bmatrix}
\sigma_{w}^2I_{n}&0\\
0&\sigma_{u}^2I_{p}\end{bmatrix}\succeq 
\sigma_{min}^2I_{n+p}.
\end{aligned}
\end{equation*}
Consequently, after some algebraic manipulations, we can apply Lemma \ref{lemma:PE} to obtain with probability at least $1-\bar{\delta}$
\begin{equation*}
\begin{aligned}
&Z^{ref}_{k}(Z^{ref}_{k})'+\lambda I_{n+p}=\sum_{t=0}^{N-2}\bar{z}_{t}\bar{z}_{t}'+\lambda I_{n+p}\\
&\succeq(\frac{(N-2)\sigma_{min}^2}{40}+\lambda) I_{n+p}\succeq \frac{N\sigma_{min}^2+42\lambda}{42} I_{n+p},
\end{aligned}
\end{equation*}
when $N\geq \max(42,200(n+p) \log(\frac{13}{\bar{\delta}}))$. 

The result then follows by applying a union bound. 
\end{pf}

\subsection{Proof of Lemma \ref{markov}}
\begin{pf}
Recall that we have
\begin{equation}  
\|Z_{k^*+N-2}^{ref}(Z_{k^*+N-2}^{ref})'\|=\|\sum_{t=k^*-N}^{k^*-2}z_{t}z_{t}'\|,
\end{equation} and
\begin{equation} \label{touse2}
\|Z_{k^*+N-2}^{test}(Z_{k^*+N-2}^{test})'\|=\|\sum_{t=k^*}^{k^*+N-2}z_{t}z_{t}'\|.
\end{equation}
We will only show the first inequality, as the second one is almost identical. From the Markov inequality, we have with probability at least $1-\bar{\delta}$
\begin{equation}
\|Z_{k^*+N-2}^{ref}(Z_{k^*+N-2}^{ref})'\|\leq \frac{\mathbf{E}[\|\sum_{t=k^*-N}^{k^*-2}z_{t}z_{t}'\|]}{\bar{\delta}}.
\end{equation}
Now we bound the term $\mathbf{E}[\|\sum_{t=k^*-N}^{k^*-2}z_{t}z_{t}'\|]$. Since the term $z_{t}z_{t}'$ has unit rank, we have $\|z_{t}z_{t}'\|=\tr(z_{t}z_{t}')$, and hence
\begin{equation}
\begin{aligned}
\mathbf{E}[\|\sum_{t=k^*-N}^{k^*-2}z_{t}z_{t}'\|]&\leq \sum_{t=k^*-N}^{k^*-2}\mathbf{E}[\|z_{t}z_{t}'\|]\\
&=\sum_{t=k^*-N}^{k^*-2} \tr(\mathbf{E}[z_{t}z_{t}']).
\end{aligned}
\end{equation}
Note that for all $t\geq 0$, from Assumption \ref{boundedV}, we have
\begin{equation}
\begin{aligned}
&\tr(\mathbf{E}[z_{t}z_{t}'])=\tr\begin{bmatrix}
\mathbf{E}[x_{t}x_{t}']&0\\
0&\sigma_{u}^{2}I_{p}
\end{bmatrix}\leq \beta+\sigma_{u}^2 p.
\end{aligned}
\end{equation}
Hence, we have with probability at least $1-\bar{\delta}$
\begin{equation}
\begin{aligned}
\|Z_{k^*+N-2}^{ref}(Z_{k^*+N-2}^{ref})'\|&\leq \frac{(N-1)(\beta+\sigma_{u}^2 p)}{\bar{\delta}}.
\end{aligned}
\end{equation}

Following a similar procedure for \eqref{touse2} and applying a union bound, we have the desired result.
\end{pf}

\bibliographystyle{plain}        

 




\end{document}